\pdfoutput=1 

\documentclass[11pt]{article}
\usepackage[final]{acl}
\usepackage{times}
\usepackage{colortbl}
\usepackage{booktabs}
\usepackage{multirow}
\usepackage{inconsolata}
\usepackage{bm}
\usepackage{arydshln}
\usepackage{latexsym}
\usepackage{wrapfig}
\usepackage{arydshln}
\usepackage{mathtools}
\usepackage{amsthm}
\usepackage{amsmath} 
\usepackage{amsmath, amssymb}
\usepackage{float} 
\usepackage{enumitem}
\usepackage[T1]{fontenc}

\usepackage[utf8]{inputenc} 

\usepackage{microtype}

\usepackage{inconsolata}
\usepackage{enumitem}
\usepackage{graphicx}

\title{Layer-Specific Scaling of Positional Encodings \\ for Superior Long-Context Modeling}

\author{Zhenghua Wang\thanks{Equal Contribution}$^{1,3}$, \
{\bf Yiran Ding}\footnotemark[1]$^{2}$, \
{\bf Changze Lv}\footnotemark[1]$^{1,3}$, \
{\bf Zhibo Xu}$^{1,3}$, \
{\bf Tianlong Li}$^{1,3}$, \\
{\bf Tianyuan Shi}$^{1,3}$, \
{\bf Xiaoqing Zheng\thanks{Corresponding Author}}$^{1,3}$, \
{\bf Xuanjing Huang}$^{1,3}$ \\
$^{1}$Fudan University \quad $^{2}$Westlake University\\
$^{3}$Shanghai Key Laboratory of Intelligent Information Processing \\
\texttt{$\{$zhenghuawang23,czlv24$\}$@m.fudan.edu.cn},
\texttt{$\{$yiran.ding$\}$@hdu.edu.cn}\\
\texttt{$\{$zhengxq,xjhuang$\}$@fudan.edu.cn}
}

\definecolor{lightpurple}{rgb}{0.9137, 0.9255, 0.9647}
\definecolor{darkpurple}{rgb}{0.8118, 0.8392, 0.9255}
\definecolor{lightgreen}{rgb}{0.8902, 0.9490, 0.8510}
\definecolor{darkgreen}{rgb}{0.7843, 0.8980, 0.7020}

\begin{document}

\maketitle
\begin{abstract}

Although large language models (LLMs) have achieved significant progress in handling long-context inputs, they still suffer from the ``lost-in-the-middle’’ problem, where crucial information in the middle of the context is often underrepresented or lost. Our extensive experiments reveal that this issue may arise from the rapid long-term decay in Rotary Position Embedding (RoPE). 
To address this problem, we propose a layer-specific positional encoding scaling method that assigns distinct scaling factors to each layer, slowing down the decay rate caused by RoPE to make the model pay more attention to the middle context. A specially designed genetic algorithm is employed to efficiently select the optimal scaling factors for each layer by incorporating Bézier curves to reduce the search space.
Through comprehensive experimentation, we demonstrate that our method significantly alleviates the ``lost-in-the-middle'' problem. Our approach results in an average accuracy improvement of up to $20\%$ on the Key-Value Retrieval dataset. Furthermore, we show that layer-specific interpolation, as opposed to uniform interpolation across all layers, enhances the model's extrapolation capabilities when combined with PI and Dynamic-NTK positional encoding schemes.

\end{abstract}

\section{Introduction}


Recent advancements in long-context LLMs \cite{team2024gemini, dubey2024llama, liu2024deepseek} have attracted considerable attention, enabling these models to process longer inputs and tackle more complex tasks, such as long-text summarization \cite{feng2021survey, zhang2021summ}, long-context question-answering \cite{li2024long}, and advanced code generation \cite{zheng2023codegeex, liu2024your}. However, as the context length increases, these models encounter the ``lost-in-the-middle'' problem \cite{liu2024lost, li2023loogle}, where they disproportionately focus on the beginning and end of the context while overlooking critical information in the middle. This issue can substantially degrade performance across various long-context tasks.

\begin{figure}[t]
\centering
\begin{minipage}{0.47\textwidth}
\centering
\includegraphics[width=\linewidth]{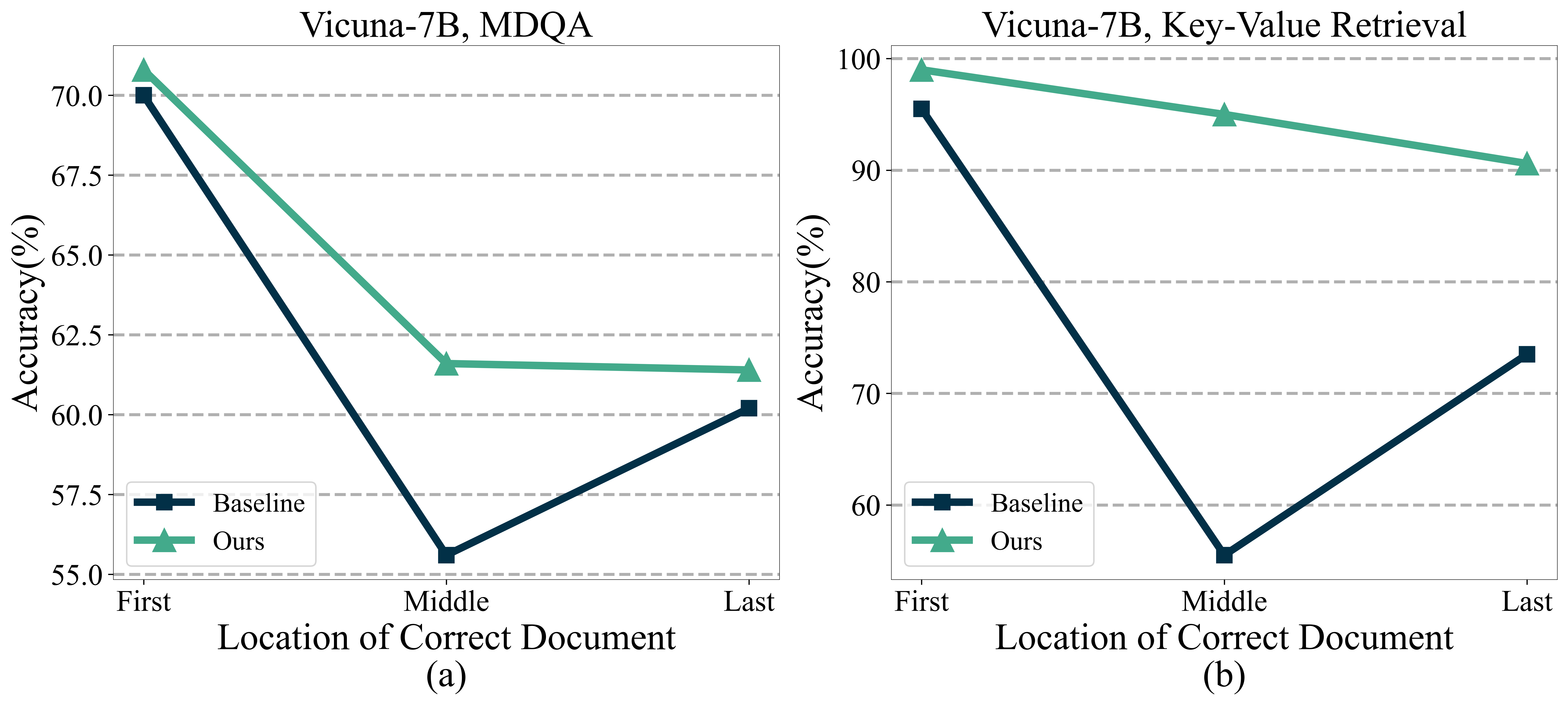}
\label{fig:first}
\end{minipage}
\vspace{-12pt}
\caption{
Average accuracy on MDQA (a) and the Key-Value Retrieval (b) datasets. By applying layer-specific scaling to enhance middle-context attention, we achieved an average accuracy improvement of +$20\%$ on the Key-Value Retrieval dataset and +$2.7\%$ on the MDQA dataset.
}
\label{Intro: two_cases}
\vspace{-4mm}
\end{figure}

RoPE is a widely used positional encoding in Transformer-based LLMs, designed to capture relative distances between input tokens while gradually reducing inter-token dependencies as the distance increases  \cite{su2024roformer}. 
Recent studies suggest that this long-term decay effect may contribute to the ``lost-in-the-middle'' problem. To address this, \citet{zhang2024found} proposes assigning different scaling factors to attention heads based on their sensitivity to relevant information. However, this approach has three key shortcomings:


\begin{enumerate}[left=0pt, topsep=0.2em]
\setlength{\parsep}{0pt}
\setlength{\parskip}{0pt}
\renewcommand{\labelenumi}{\Roman{enumi}.}
\item    
Although scaling factors are dynamically assigned based on the properties of each attention head, they are applied within the same range across all layers, which may fail to account for layer-specific variations~\cite{li2024happened, men2024shortgpt, sun2024transformer}.

\item There is no systematic approach that generalizes across different models and datasets for effectively determining the optimal range of scaling factors. 

\item During inference, this method requires computing each attention head’s sensitivity to relevant information to determine its scaling factor. This incurs additional time overhead, resulting in prolonged response times. In contrast, our approach eliminates this burden, maintaining efficiency while enhancing model performance.
\end{enumerate}


\noindent Additionally, preliminary experiments indicate that a uniform scaling strategy that assigns the same scaling factors across all layers and heads can alleviate the ``lost-in-the-middle'' problem. However, it introduces a new issue: ``lost-in-the-tail'', where the model's ability to capture information at the end of input texts is weakened. To address the limitations mentioned above, we seek to assign distinct scaling factors to each layer, as our findings suggest that scaling different layers yields varying effects, as showing in Table \ref{more_part_layer}.




Determining appropriate scaling factors for each layer is challenging. 
Assuming that the appropriate scaling factor for each layer is selected from a finite set of scaling factors, as each layer can have multiple choices, leading to a rapidly expanding combinatorial optimization problem as the number of layers increases. 
To address this, we apply a genetic algorithm to solve the optimization problem, limiting the search space to Bézier curves which approximately boosts search speed by a factor of \(10^{20}\) over brute-force search across \(32\) layers. A Bézier curve, defined by a small number of key points, is used to model the functional relationship between scaling factors and layer depths.
By selecting these points, we can determine the corresponding Bézier curve and, consequently, the scaling factors for each layer.


By combining the specially designed genetic algorithm with Bézier curves, we can optimize layer-specific scaling factors efficiently, typically within $4$ to $8$ hours, using a modest amount of computational resources.
For instance, the appropriate layer-specific scaling factors for the Vicuna-7B-v1.5 model \cite{Berkeley_Cmu_San} can be determined in just $6$ hours using four A100 GPUs on the MDQA dataset~\cite{Liu_Lin_Hewitt_Paranjape_Bevilacqua_Petroni_Liang}.
Given the significant improvements in long-context modeling for LLMs, investing a few hours and some computational resources to select the optimal layer-specific scaling factors is a worthwhile trade-off.

The contributions of this study can be summarized as follows.

\begin{itemize}[topsep=0.2em]
\setlength{\itemsep}{0pt}
\setlength{\parsep}{0pt}
\setlength{\parskip}{0pt}

\item We propose a layer-specific positional encoding scaling method that effectively alleviates the ``lost-in-the-middle'' problem, thereby enhancing the model's ability to handle long-context tasks. Furthermore, this approach is highly versatile and can be applied to a wide range of models.

\item We propose an efficient and fast genetic search algorithm, where the search space is constrained by Bézier curves, enabling the rapid determination of the scaling factors for each layer using only a few representative examples.

\item We investigate the underlying causes of the ``lost-in-the-middle'' problem and offer a reasonable explanation for why our layer-specific positional encoding scaling method can effectively mitigate this problem.

\end{itemize}




\section{Related Work}

\subsection{Positional Encoding}

Positional encoding can be categorized into two main types: absolute positional encoding \cite{Clark_Luong_Le_Manning_2020,Lan_Chen_Goodman_Gimpel_Sharma_Soricut_2019} and relative positional encoding \cite{Su_Lu_Pan_Wen_Liu_2021,huang2020improve,shaw2018self}. Absolute positional encoding directly adds the absolute positional information to token embedding, while relative positional encoding models the relative distance between tokens. In relative positional encoding, RoPE has become a mainstream method due to its excellent extrapolation property and adaptability to linear attention.

With RoPE \cite{Su_Lu_Pan_Wen_Liu_2021}, the relative positional relationships are reflected through the dot product operation: 
\begin{equation}
f\left(\mathbf{q}_{m}, m\right)^{T} f\left(\mathbf{k}_{n}, n\right)=g\left(\mathbf{q}_{m}, \mathbf{k}_{n}, m-n\right)
\end{equation}
The function $f$ represents the rotation of the query vector $q$ at position  $m$ and the key vector $k$ at position $n$, incorporating positional information into them. The dot product of the rotated vectors reflects the relative position \( m-n\).

Position interpolation (PI) \cite{chen2023extending} expands the context window by narrowing the interval between position indices. The RoPE function \(f\) is replaced by \(f'\) defined as follows:
\begin{equation}
\begin{aligned}
    f'(x,m) &= f(x,\frac{m}{s}) \\
    s &= \frac{L'}{L}
\end{aligned}
\end{equation}
\noindent where \(L\) is pretrained  context window size of the LLMs and \(L'\) is the extended context window size, and  \(m\) is the position index. PI scales the original relative distance by a factor of \(\frac{1}{s}\).


\subsection{The Lost-in-the-Middle Phenomenon}

The ``lost-in-the-middle'' issue is a prevalent challenge in both open-source and closed-source models 
\cite{liu2024lost}. This issue arises when critical information is positioned in the middle of the model's context, causing the model to struggle with effectively processing and understanding the content, which ultimately results in a performance decline. Recent approaches to mitigate this problem can be broadly categorized into three types: fine-tuning-based methods, positional encoding-based methods, and attention-based methods.

In fine-tuning-based methods, \citet{he2024never,an2024make} propose constructing a dataset that emphasizes information located in the middle positions. By applying instruction fine-tuning, the model can be guided to pay more attention to the central part of the context. However, this approach requires considerable human effort to curate the relevant samples and entails substantial computational costs for the fine-tuning process. In positional encoding-based methods, the primary approach is to interpolate RoPE to mitigate long-term decay, enabling the model to better focus on information in the middle of the context. \citet{zhang2024found} performs interpolation at the head level of multi-head attention, while \citet{chen2023fortify} applies interpolation at the model level, where multiple models with different scaling factors run in parallel and decode the next token based on confidence scores. However, both methods introduce substantial time and memory overhead. In attention-based methods, \citet{peysakhovich2023attention} reposition relevant documents to areas of the context where the model's attention is most concentrated, based on attention weights. \citet{hsieh2024found} formulates a positional bias equation, rescaling the attention score matrix to address the model's bias toward specific positions.


\subsection{Bézier Curve}

Bézier curves are parametric curves widely utilized in computer graphics and related fields \cite{mortenson1999mathematics}. They are defined by a set of control points, which generate a smooth and continuous curve through a mathematical formula. The defining feature of Bézier curves is that a small number of control points can precisely shape complex curves, making them especially valuable in applications such as automotive body design \cite{mineur1998shape,zheng2020bezier}, font creation \cite{haryono2014studi}, and animation \cite{bargteil2014animation}

A Bézier curve of degree \( n \) is defined as:
\begin{equation}
    \small
    B(t) = \sum_{i=0}^{n} B_i P_i
\end{equation}
where \( P_i \) are the control points, \( B_i \) are the Bernstein basis polynomials given by:
\begin{equation}
    \small
    B_i^n(t) = \binom{n}{i} (1 - t)^{n-i} t^i, \quad t \in [0,1]
\end{equation}
where \( \binom{n}{i} \) is the \textbf{binomial coefficient}:
\begin{equation}
    \binom{n}{i} = \frac{n!}{i!(n-i)!}
\end{equation}

\section{Method}

\begin{figure*}[h]
\centering
\includegraphics[width=0.85\linewidth]{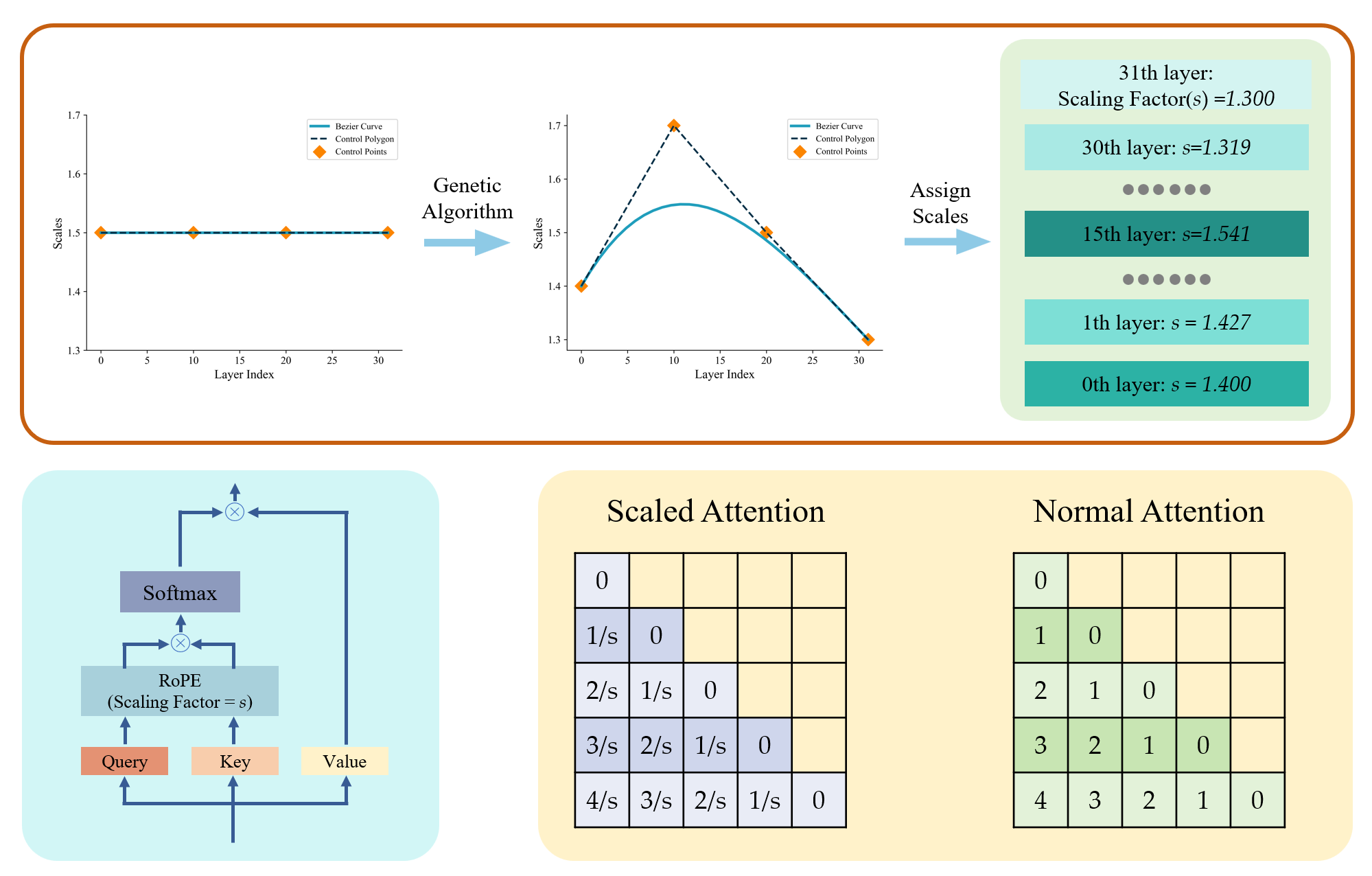}
\caption{
\textbf{Modeling with Bézier Curves}: First, we optimize the initial control points to optimal points using a constrained genetic algorithm. Then, we apply scaling factors derived from the fitted curve to each layer. The bottom part of the figure illustrates the model structure, alongside a comparison between the scaled attention and the normal attention.
}
\label{Method: work_flow}
\end{figure*}

In this section, we first conduct experiments to validate the potential causes of the ``lost-in-the-middle'' phenomenon and establish the rationale for layer-specific positional encoding scaling(\textsection \ref{method: Layer-wise Scaling Enhanced Context Utilization}). Next, we formally define the problem using mathematical formulations, aiming to enhance the model's ability to utilize middle-position information without degrading its original performance(\textsection \ref{method: Problem Formulation}). Finally, we present a detailed methodology for assigning optimal scaling factors to each layer, leveraging a combination of genetic algorithms and Bézier curves(\textsection \ref{method: Searching the Layer-wise Position Interpolation scales}). The overall workflow is represented in Figure \ref{Method: work_flow}.

\subsection{Layer-Specific Scaling Enhanced Context Utilization}
\label{method: Layer-wise Scaling Enhanced Context Utilization}

\begin{figure}[h]
    \centering
    \includegraphics[width=\linewidth]{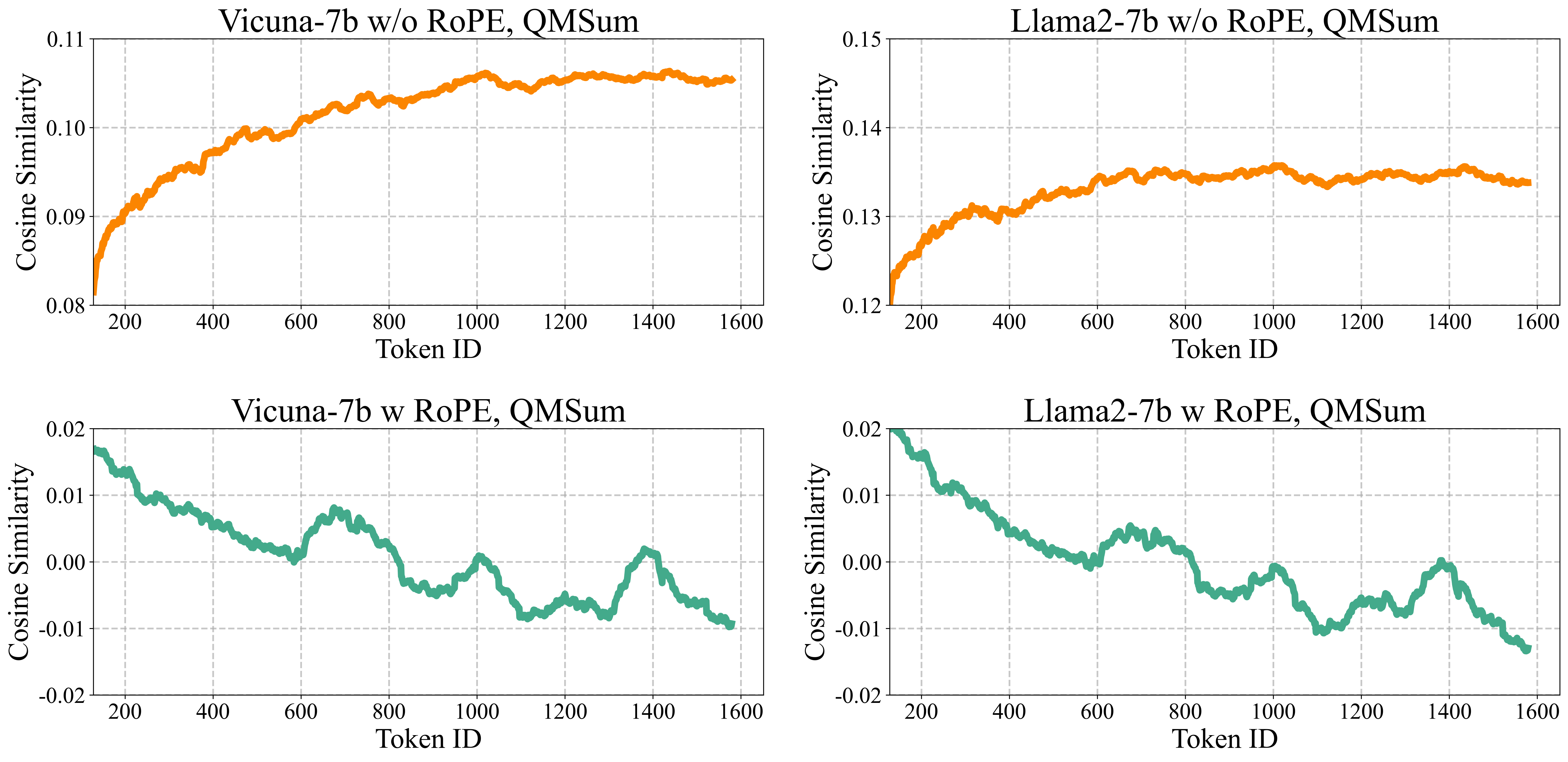}
    \caption{
    Cosine similarity between the average of first 128 tokens and the later tokens reveals the model heightened focus on earlier content without positional encoding.
    }
    \label{causal_attention_effect}
    \vspace{-3mm}
\end{figure}

The ``lost-in-the-middle'' phenomenon causes the model to focus more on the ends of the context while neglecting the middle information, which may be due to the following two reasons: 

\noindent\textbf{(1) Causal Attention Mechanism}: The causal attention mechanism causes the model to focus more on the beginning of the sequence, as earlier tokens are involved in more attention calculations. To investigate the effect of causal attention, we use the \textbf{Vicuna-7B-v1.5} \cite{Berkeley_Cmu_San} and \textbf{LLaMA-2-7B-hf} \cite{touvron2023llama} to conduct experiment on the \textbf{QMSum} dataset \cite{Shaham_Ivgi_Efrat_Berant_Levy_2023}. We randomly sample \(50\) items from the dataset and remove the RoPE from layers \(2\) to \(4\). Then, we extract the hidden states from layer \(4\) and calculate the average cosine similarity between the mean representations of the first \(128\) tokens and representations of the subsequent tokens. As shown in Figure \ref{causal_attention_effect}, the representations of later tokens become more similar to those of earlier tokens in the absence of RoPE, which is the opposite of the normal trend. This demonstrates that the model's causal attention makes it focus more on earlier content.

\begin{figure}[ht]
    \centering
    \includegraphics[width=\linewidth]{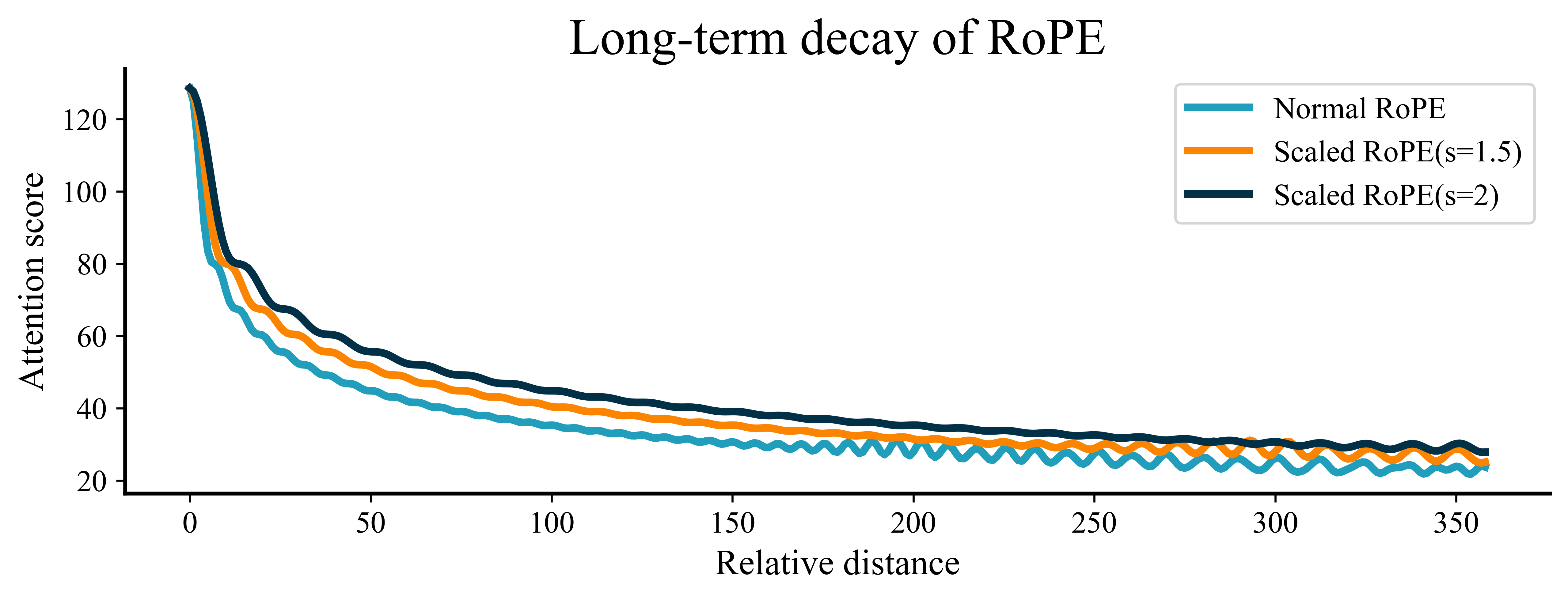}
    \caption{
    The rapid decay of RoPE biases local focus, while the scaling operation can slow the decay, contributing to the enhancement of global attention capacity.
    }
    \label{long-term-decay}
\end{figure}

\begin{figure}[h]
    \centering
    \includegraphics[width=\linewidth]{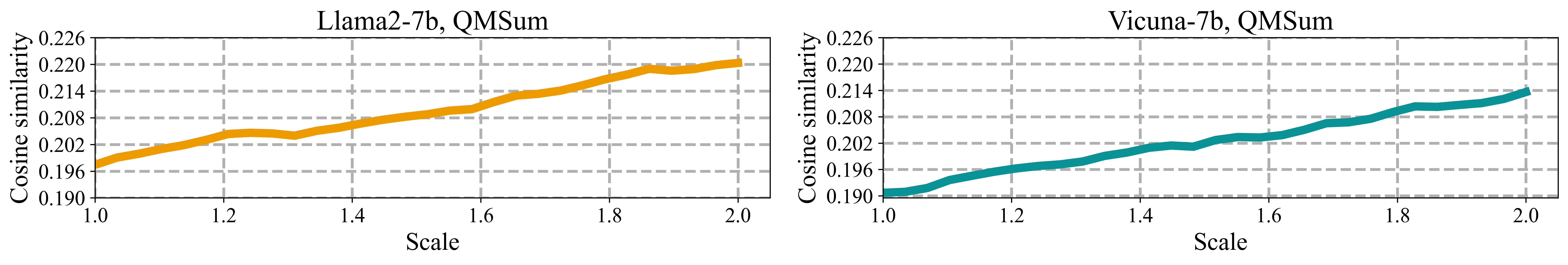}
    \caption{
    Cosine similarity between the average of middle context and the last token reveals the model heightened focus on middle context as the scaling factor increases.
    }
    %
    \label{scale_cos}
\end{figure}

\noindent \textbf{(2) Long-term decay of RoPE}: The Long-term decay of RoPE causes the model to focus more on the end of the sequence. As the relative distance increases, the attention scores decay rapidly, leading the model to focus excessively on nearby tokens during autoregressive decoding, while neglecting information from distant tokens. 
We scale the RoPE by a factor of \(s\), reducing the relative distance to \(1/s\) of its original value. This alleviates the decay rate as shown in Figure \ref{long-term-decay}, allowing the model to focus not only on nearby tokens but also on more distant tokens, particularly those in the middle positions. Adopting the experimental setup described previously, additionally, we partitioned the context into three parts and calculated the cosine similarity between the average representation of the middle part tokens and the representation of the last token at different scales. As the scale increases, the similarity between representations improves which is shown in Figure \ref{scale_cos}, demonstrating that scaling RoPE allows the model to focus more on the content of the middle part during autoregressive decoding. 
 
To mitigate the ``lost-in-the-middle'' problem, we have two options: shift excessive attention from the start of the context to the middle positions or redistribute attention from the end of the context. Given that causal attention is a fundamental part of the model's architecture and is difficult to modify, we opted for the latter. As depicted in the bottom of Figure \ref{Method: work_flow} and in Figure \ref{long-term-decay}, we redistribute attention from the later part to the middle part by scaling RoPE.

Applying uniformly large scaling factors to layers can ease the ``lost-in-the-middle'' problem, yet it may neglect the content at the end position \cite{zhang2024found}. We suggest applying scaling factors to specific layers, as shown in Table \ref{more_part_layer}, to enhance the model's utilization of the end part context. Our findings indicate that scaling earlier layers likely improves the model's comprehension of the context's ending part while scaling later layers tends to enhance its understanding of the context's beginning part. Therefore, we should dynamically assign scaling factors to layers based on their characteristics.
%
%
%
%

%
%
\subsection{Problem Formulation} \label{method: Problem Formulation}
To efficiently determine the optimal scaling factors for each layer, we employ the Bézier curve to model the functional relationship between layer depths and scaling factors. Subsequently, a genetic algorithm is applied to determine the control points of the curve.
Given the model parameters \( \theta \) and the dataset \( X \), our goal is to boost the model's exploitation of context at middle positions while ensuring that the performance at both ends of the context does not degrade. This goal can be formulated as follows:
\begin{equation}
\underset{s_i \in \mathbf{S} ; s_i \geq 1}{\arg \min } \mathbb{E} \left[ \sum\limits_{x \in \mathbf{X}} \mathcal{L}(x,\operatorname{LLM}(\operatorname{\theta}, \mathbf{S}))\right]
\end{equation}
\(S\) is the set of scaling factors for each layer of the model. To slow the long-term decay phenomenon and prevent harm to the model's context window, each element \( s_i\) in the set is greater than \(1\), as shown in Figure \ref{long-term-decay}. \textbf{LLM} is a function that scales the positional encoding of the model's layers according to the set S.
\begin{equation}
    \mathbf{S} = B(t), \quad t \in \left\{ \frac{k}{n} \mid k = 0, 1, 2, \dots, n \right\}
\end{equation}
\(n\) is the number of layers that will be scaled. Since the value of \(t\) ranges from \(0\) to \(1\), we divide 
\(t\) into equal intervals according to the number of layers.
\subsection{Searching the Layer-Specific Scaling Factors} \label{method: Searching the Layer-wise Position Interpolation scales}
Brute-force search is impractical for determining the scaling factors for each layer due to its complex search space. Therefore, we employ a genetic algorithm to determine the control points of Bézier curves.

\noindent\textbf{Search space}: We follow \citet{ding2024longRoPE}, discretizing the continuous search space to enable more efficient searching. Assume the control points of the Bézier curve are \( (B_x, B_y) \), where \( B_x \in [0, n-1] \)(n is the number of layers) and \( B_y \in [1, 2] \). The values of \( B_x \) are discretized with a step size of \(1\), and the values of \( B_y \) are discretized with a step size of \(0.1\) which is represented in table \ref{search_space}. Assuming the model consists of \(32\) layers and the scaling factor for each layer is determined from the \(B_Y\) set, the total number of choices for a brute-force search is \(11^{32}\). If a Cubic Bezier curve is used, each control point has \(32*11\) possible combinations. With four control points, the total search space is \(352^4\) which approximately narrows the search space by a significant factor \(10^{20}\) compared to brute-force search.
\begin{table}[h]
\centering
\small
\resizebox{\linewidth}{!}{
\begin{tabular}{|c|c|}
\hline
\textbf{} & \textbf{Search Space} \\ \hline
\(B_X\) & \{$0$, $1$, $2$, $3$, $4$, $5$, $6$, $7$, $8$,$\dots$, $n-4$, $n-3$, $n-2$, $n-1$\} \\ \hline
\(B_Y\) & \{$1.0$, $1.1$, $1.2$, $1.3$, $1.4$, $1.5$, $1.6$, $1.7$, $1.8$, $1.9$, $2.0$\} \\ \hline
\end{tabular}
}
\caption{Search space for the control point of Bézier curves.}
\label{search_space}
\end{table}

\noindent\textbf{Genetic algorithm}:
We use a Cubic Bezier curve with four control points to conduct the following experimental analysis. The first individual is initialized as follows:
\begin{equation}
  \mathbf{P} =\left[ \left(\frac{(n-1) \cdot i}{3}, 1.5\right) \mid i = 0, 1, 2, 3 \right],
\end{equation}
where $n$ is the number of scaling layers, based on the experimental setup of \citet{zhang2024found}, we set the \(B_y\) of all control points to 1.5. The remaining individuals from the population will undergo mutations based on the first individual.

To ensure the stability and effectiveness of the search algorithm, the following constraints were applied to the mutation and crossover operations: 

\noindent\textbf{(1)} The \(B_x\) of all control points are increasing monotonically to ensure the smoothness of the curve to prevent abrupt changes in scaling factors. Let \( B_x^j \) and \( B_x^i \) denote the first dimension of the \( j \)-th and \( i \)-th control points, respectively. The relationship between them is as follows:
\begin{equation}
\label{constrain1}
0 \leq B_x^j < B_x^i \leq n-1 \quad if  \quad i > j 
\end{equation}
\noindent\textbf{(2)} the mutation operation modifies \(B_x\) and \(B_y\) within a specific range to prevent excessive curve variations to avoid affecting the convergence stability of the search.  Let \( N_x \) denote the amplitude of change in the first dimension(layer index), and \( N_y \) denote the amplitude of change in the second dimension(Scaling factor), \( n \) represent the number of layers. 
%
%
%
\begin{equation}
    \tiny
    \label{constrain2}
    \begin{aligned}
        \hat{B}_x^i &= 
        \begin{cases}
            [\min(0, B_x^{i} - N_x), 
                \max(B_x^{i+1}, B_x^{i} + N_x)] 
                & \text{if } i = 0, \\
            [\min(B_x^{i-1}, B_x^{i} - N_x), 
                \max(B_x^{i+1}, B_x^{i} + N_x)] 
                & \text{if } i \in (0,k), \\
            [\min(B_x^{i-1}, B_x^{i} - N_x), 
                \max(B_x^{i} + N_x, n - 1)] 
                & \text{if } i = k.
        \end{cases} \\
        \hat{B}_y^i &= 
        \begin{cases}
            [\min(1, B_y^{i} - N_y), 
                \max(B_y^{i} + N_y, 2)] 
                & \text{if } i = 0, \\
            [\min(1, B_y^{i} - N_y), 
                \max(B_y^{i} + N_y, 2)] 
                & \text{if } i \in (0,k), \\
            [\min(1, B_y^{i} - N_y), 
                \max(B_y^{i} + N_y, 2)] 
                & \text{if } i = k.
        \end{cases}
    \end{aligned}
\end{equation}
\begin{algorithm}[t]
	\small
	\caption{The search algorithm for determining layer-specific scaling factors}
	\textbf{Input:} target LLM, input samples $\mathbf{X}$,   population size $P$, mutation size $N_1$, crossover size $N_2$, max iterations $\mathcal{T}$, mutate probability $p$\\
	\vspace{-2.5ex}

	\begin{algorithmic}[1]
		\label{alg:search}
		\STATE Top-k=$\phi$;	
		\STATE $\text{P}_0$=\textit{Initialize\_population} ($P$, $\mathbf{X}$, $p$); 
		\FOR{$i$=$1$ to $\mathcal{T}$}
            \STATE \textit{Compute\_context\_utilization} (LLM, $\text{P}_{i-1}$, $\mathbf{X}$);    
		\STATE  Top-k = \textit{Update\_Topk} (Top-k, $\text{P}_{i-1}$);
		\STATE$\text{P}_{mutation}$=\textit{Mutation\_with\_constraint} (Top-k, $N_1$, $p$); 
		\STATE $\text{P}_{crossover}$=\textit{Crossover\_with\_constraint} (Top-k, $N_2$);
		\STATE $\text{P}_i$ = $\text{P}_{mutation}$ $\cup$ $\text{P}_{crossover}$ $\cup$ Top-k;
		\ENDFOR
		\STATE Return the individual with highest context utilization in Top-k;
	\end{algorithmic}
    \label{search algorithm}
\end{algorithm}

Algorithm~\ref{search algorithm} formalizes the multi-objective optimization for layer-adaptive scaling factors. The function \textbf{\textit{Compute\_context\_utilization}} is defined as a  $\mathcal{U}$. 
\begin{equation}
\mathcal{U} = \lambda_f A_f + \lambda_m A_m + \lambda_l A_l
\end{equation}
where $A_f, A_m, A_l$ denote accuracy when target documents appear in first/middle/last context positions, with constraint weights $\lambda_f, \lambda_m, \lambda_l \in \mathbb{R}^+$. To prevent degradation of last context utilization, we enforce monotonic weight ordering:
\begin{equation}
\lambda_f < \lambda_m < \lambda_l
\end{equation}
The operation \textbf{\textit{Crossover\_with\_constraint}} will swap the control points between two individuals with a certain probability. However, any changes made during the crossover process must satisfy the constrain Equation (\ref{constrain1}). If the modified crossover does not meet the constrain Equation (\ref{constrain1}), a new pair of individuals will be selected again for crossover. The maximum retry number of crossover is set  \(2N_2\), in order to prevent situations where no valid crossover individuals are found.

\section{Experiment}

In this section, we demonstrate that the layer-specific positional encoding scaling effectively mitigates the ``lost-in-the-middle'' problem and enhances the model's ability to utilize contextual information(\textsection \ref{exp: Enhanced ability to utilize contextual information.}). Moreover, compared to uniform scaled positional encoding , this method demonstrates superior performance in enhancing the model's extrapolation capability(\textsection \ref{exp: Enhanced long texts modeling ability of the model.}).
%
%
%
%
%
\subsection{Enhanced Ability to Utilize Contextual Information} \label{exp: Enhanced ability to utilize contextual information.}
\textbf{Experimental Setup} We selected three mainstream RoPE-based LLMs: Vicuna-7B-v1.5 \cite{Berkeley_Cmu_San}, LLaMA-2-7B-chat \cite{touvron2023llama}, and StableBeluga-7B \cite{StableBelugaModels} for our experiments. These models all have 32 layers, and in the experiment, we only scale the last 30 layers, followed \citet{zhang2024found}.
\begin{table}[H]
\centering
\small
\resizebox{\linewidth}{!}{
\begin{tabular}{l|l|ccc|c}
\hline
\bf Models & \bf Methods & \bf First & \bf Middle & \bf Last & \bf Avg.  \\
\hline
\multicolumn{6}{c}{\textit{MDQA}} \\ \hline

\multirow{4}{*}{Vicuna-7B-v1.5} & Baseline & 70.0  & 55.6  & 60.2  & 61.9  \\ 
& PI & 70.6  & 59.6  & 59.8  & 63.3  \\ 
& Ms-PoE & 71.8  & 62.8  & 60.0  & 64.8  \\ 
& Ours & \cellcolor{darkgreen}70.8  & \cellcolor{darkgreen}61.6  & \cellcolor{darkgreen}61.4  & \cellcolor{darkgreen}64.6   \\ \hline

\multirow{4}{*}{StableBeluga-7B} & Baseline & 68.8  & 58.8  & 70.8  & 66.1  \\ 
& PI & 72.0  & 58.8  & 69.4  & 66.7 \\
& Ms-PoE & 73.0  & 59.4  & 64.0  & 65.5  \\ 
& Ours & \cellcolor{darkgreen}72.0  & \cellcolor{darkgreen}62.0  & \cellcolor{darkgreen}71.0  & \cellcolor{darkgreen}68.3  \\ \hline

\multicolumn{6}{c}{\textit{Key-Value Retrieval}} \\ \hline
\multirow{4}{*}{Vicuna-7B-v1.5} & Baseline & 95.5  & 55.5  & 73.5  & 74.8  \\ 
& PI & 98.0  & 90.5  & 84.0  & 90.8  \\ 
& Ms-PoE & 97.5  & 69.5  & 72.5  & 79.8  \\ 
& Ours & \cellcolor{darkgreen}99.0  & \cellcolor{darkgreen}95.0  & \cellcolor{darkgreen}90.5  & \cellcolor{darkgreen}94.8   \\ \hline

\multirow{4}{*}{StableBeluga-7B} & Baseline & 90.5  & 6.00  & 81.50  & 51.0  \\ 
& PI & 94.0  & 22.5  & 83.5  & 66.1 \\
& Ms-PoE & 89.5  & 34.0  & 77.5  & 67.0 \\
& Ours & \cellcolor{darkgreen}98.5  & \cellcolor{darkgreen}37.0  & \cellcolor{darkgreen}87.0  & \cellcolor{darkgreen}68.33  \\ \hline
\end{tabular}
}
\caption{``First'', ``Middle'', and ``Last'' indicate that the correct document is located at the beginning, middle, and end of the context. 
}
\label{MDQA_kv}
\end{table}

MDQA \cite{Liu_Lin_Hewitt_Paranjape_Bevilacqua_Petroni_Liang} is a multi-document question-answering dataset. Due to the context window limitations of LLMs, we selected 10 documents as context and evaluated the model’s answer accuracy when the correct document appeared at different positions. Key-Value Retrieval dataset \cite{Liu_Lin_Hewitt_Paranjape_Bevilacqua_Petroni_Liang} consists of key-value pairs where both keys and values are UUIDs, enabling an assessment of the model's retrieval capability across different positions without semantic interference. ZeroSCROLLS \cite{Shaham_Ivgi_Efrat_Berant_Levy_2023} includes multiple long-text task datasets. we selected seven sub-datasets: GovReport, SummScreenFD, Qasper, QMSum, NarrativeQA, SQuALITY, and SpaceDigest to evaluate the model’s performance in summarization, question answering, and comprehension under long-context scenarios. We set the context window to \(3,500\) tokens and limit the maximum number of decoded tokens to \(500\).

In table \ref{MDQA_kv}, PI represents the scaling factor for each layer is the same, and its value is the average of the layer-specific scaling factors. Ms-PoE dynamically assigns scale factors ranging from 1.2 to h1.8 to the heads based on their sensitivity to relevant information across all layers. The more sensitive an attention head is to relevant information, the smaller the scaling factor assigned to it.

\noindent\textbf{Result}: 
We primarily discuss our experimental results from three aspects: performance, generalization, and efficiency. 

\begin{enumerate}[left=0pt, topsep=0.2em]
\setlength{\parsep}{0pt}
\setlength{\parskip}{0pt}
\renewcommand{\labelenumi}{\Roman{enumi}.}
\item \textbf{Dynamic scaling factor allocation based on layer characteristics enhances contextual utilization:} As shown in Table \ref{MDQA_kv}, our method mitigates the ``lost-in-the-middle'' phenomenon while maintaining performance at context both ends. Compared to baselines, it achieves consistent gains, most notably a 20.0-point improvement on Key-Value Retrieval. In contrast, uniform scaling across layers (e.g., PI and Ms-PoE) improves mid-context utilization but impairs information retrieval from context both ends. On MDQA, both PI and Ms-PoE degrade performance at end positions, underscoring the necessity of layer-specific scaling.

Across multiple long-text tasks of Table \ref{zeroscroll} in the Appendix, all three models showed overall improvements, ranging from \(1.66\) to \(2.44\). Notably, Vicuna-7B-v1.5 demonstrated an improvement of \(10.55\) on the Qasper dataset. This demonstrates that incorporating a layer-specific scaling method can significantly enhance the model's utilization of contextual information. 

\item \textbf{Our systematic search algorithm efficiently identifies layer-specific scaling factors across models with minimal data.} While Ms-PoE \cite{zhang2024found} manually tailors scaling factors for Vicuna, this approach fails to generalize: transferred to StableBeluga-7B \cite{StableBelugaModels}, it degrades end-position performance by \(6.8\%\) on MDQA. Moreover, Ms-PoE lacks a systematic scaling factors selection framework. By contrast, our method identifies optimal layer-specific scaling factors for diverse models using only \(10\%\) data on MDQA. For instance, it determines optimal scaling factors for Vicuna-7B-v1.5 \cite{Berkeley_Cmu_San} in \(6\) hours on four A100 GPUs with MDQA, demonstrating both efficiency and generalizability.



\item \textbf{Our method does not introduce any delay to the model's inference speed.} Ms-PoE \cite{zhang2024found} requires real-time calculation of each head's sensitivity to relevant information, and different scaled positional encoding must be assigned individually to each attention head across almost all layers, which results in a decrease in inference speed. Using one 3090 GPU, we measured the average inference time per sample for the Vicuna-7B-v1.5 on the MDQA dataset. For Ms-PoE, the inference time per sample is \(4.4\) seconds, while our method requires only \(3.1\) seconds, which is roughly faster \(30\%\), making our approach more efficient.
\end{enumerate}

\subsection{Enhanced Extrapolation Ability of the Model} \label{exp: Enhanced long texts modeling ability of the model.}

\textbf{Experimental Setup}: Vicuna-7B-v1.5 \cite{Berkeley_Cmu_San} and LLaMA-2-7B-hf \cite{touvron2023llama}, both with a 4K context window, were used to conduct experiments on 100 samples extracted from the validation set of the PG19 dataset \cite{Rae_Potapenko_Jayakumar_Lillicrap_2019} to validate the effectiveness of layer-specific positional encoding scaling in improving the model's long-text extrapolation capability. Perplexity (PPL) at sequence lengths of 5K, 6K, 7K, and 8K is used to assess the extrapolation ability of model. If the extrapolation length is \( L’ \) and the model's context window size is \( L \), we define the scaling factor as 
\begin{equation}
    s = \frac{ L’ }{L}
\end{equation}
To explore the advantages of layer-specific positional encoding, we introduced an interval of \(0.3\), allowing the scale to vary between the minimum value \( s \) and the maximum value \( s + 0.3 \). As the layer index increases, the scaling factor first increases linearly to its maximum value and then decreases linearly back to its minimum value. The rationale behind this varying trend is discussed in detail in the Discussion section \ref{Discussion}.

PI \cite{chen2023extending} extends the context window by scaling position indices, and Dynamic-NTK~\cite{dynamicntk} extends it by scaling the base of positional encoding. These methods apply uniform scaling across all layers of the model, overlooking the characteristics of layers. We set the scaling factor of PI and Dynamic-NTK as the average of the layer-specific scaling factors set: \( s + 0.15\) and is applied to every layer of the model to investigate whether layer-specific positional encoding scaling can improve the performance of these methods.  

\noindent\textbf{Result}: In table \ref{longcontext_ppl}, Layer-specific scaling method achieves lower perplexity across various methods and extrapolation lengths. When the context window exceeds the pretrained length, the out-of-distribution (OOD) issue \cite{Jin_Han_Yang_Jiang_Liu_Chang_Chen_Hu_2024} arises in attention distribution across the model's layers. PI, Dynamic-NTK, and various extrapolation techniques can restore the attention distribution to normal, but the sensitivity to OOD issues varies across layers, suggesting that the scaling factor should differ for each layer. 

\begin{table}[ht]
\centering
\small
\tabcolsep=0.1cm
\resizebox{\linewidth}{!}{
\begin{tabular}{l|l|cccc}
\hline
\bf Model &  \bf Method & \bf 5k & \bf 6k & \bf 7k & \bf 8k \\ \hline
\multirow{4}{*}{LLaMA-2-7B-hf}& PI & 7.306  & 8.619  & 9.507  & 10.41  \\ 
& Ours-PI & \cellcolor{darkgreen}7.075  & \cellcolor{darkgreen}8.366  & \cellcolor{darkgreen}9.365  & \cellcolor{darkgreen}10.28  \\  
& Dy-NTK & 6.947  & 7.814  & 8.703  & 9.507  \\   
& Ours-Dy-NTK & \cellcolor{darkgreen}6.945  & \cellcolor{darkgreen}7.777  & \cellcolor{darkgreen}8.694  & \cellcolor{darkgreen}9.499  \\  \hline

 \multirow{4}{*}{Vicuna-7B-v1.5}& PI & 8.662  & 10.28  & 11.76  & 12.97 \\ 
  & Ours-PI & \cellcolor{darkgreen}8.559  & \cellcolor{darkgreen}10.03  & \cellcolor{darkgreen}11.56  & \cellcolor{darkgreen}12.87  \\ 
 & Dy-NTK & 8.467  & 9.609  & 10.67  & 11.66 \\ 
 & Ours-Dy-NTK & \cellcolor{darkgreen}8.478  & \cellcolor{darkgreen}9.598  & \cellcolor{darkgreen}10.57  & \cellcolor{darkgreen}11.61  \\ \hline 
 
\end{tabular}
}
\caption{
Perplexity of 100 samples on PG19 Validate datasets.
This further confirms that the layer-specific scaling method outperforms uniformed scaling(PI, Dy-NTK) in improving the model's extrapolation capability.
}
\label{longcontext_ppl}
\end{table}
\vspace{-12pt}
\section{Discussion} \label{Discussion}

Attention entropy reflects the degree of focus in the model’s attention and is calculated as follows:

\begin{equation}
H = -\sum_{i=1}^{n} p_i \log p_i
\end{equation}
where \( p_i \) is the normalized attention score for the \( i \)-th token. A smaller entropy value indicates that the model's attention is more focused.

In Figure \ref{disscussion_entropy}, when the model extrapolates beyond the pre-trained context window size, we observe a significant increase in the model's attention entropy, as shown in  Figure \ref{disscussion_entropy}(a). Through interpolation, the attention entropy is reduced and approaches a normal distribution, as shown in  Figure \ref{disscussion_entropy}(b). 
However, Whether within or beyond the pretraining window, as the scaling factor increases, the attention entropy first \textit{rises and then falls}, reflecting a shift from concentrated to more dispersed attention, as shown in  Figure \ref{disscussion_entropy}(b), (c).

As shown in Figure \ref{B_disscussion_mdqa} and Figure \ref{B_disscussion_KV} in the Appendix, the scaling factors for the earlier and middle layers are relatively large, while the scaling factors for the later layers are smaller. Larger scaling factors tend to disperse the attention across more tokens, allowing the model to capture a broader context, while smaller scaling factors help to concentrate the attention on more relevant tokens, filtering out irrelevant information. Based on this observation, we hypothesize that the earlier layers play a key role in aggregating information, while the later layers specialize in filtering and refining the most pertinent information. We further suggest that larger scales provide the model with a broader perspective, while smaller scales enable the model to focus more sharply on the information most relevant to the query, minimizing interference from irrelevant data.

Additionally, our analysis revealed that attention entropy in middle layers is comparatively lower than in the model’s initial and final layers. This reduced entropy suggests that middle-layer attention mechanisms exhibit heightened sensitivity to entropy fluctuations during extrapolation. We hypothesize that strategically assigning larger scaling factors to these middle layers during interpolation could strengthen the model’s extrapolation capabilities by stabilizing their attention variation. To validate these insights, we intend to pursue rigorous empirical studies and theoretical analyses in subsequent research.


\vspace{-2mm}
\section{Conclusion}
\vspace{-3mm}

In this paper, we addressed the ``lost-in-the-middle'' problem that has hindered the performance of LLMs in long-context tasks. Through our proposed layer-specific positional encoding scaling method, we demonstrated a novel approach that alleviates this issue by assigning distinct scaling factors to each layer, effectively slowing down the long-term decay rate of RoPE. We also introduced an efficient genetic algorithm, constrained by Bézier curves, to determine the optimal scaling factors with minimal computational resources.
This study not only advances the state of long-context modeling but also offers a practical solution that can be readily applied to various large language models. 

\section*{Limitations}

Due to computational resource constraints, we did not conduct experiments on models larger than 7B. However, the consistent performance improvements observed across multiple 7B models confirm the effectiveness of our method.

Although some current LLMs do not have serious ``lost-in-the-middle'' issue, due to their tricks on post-training with a lot data and compute resources, which is orthogonal to our method. Thus, we do not explore these models in this paper.

Additionally, given the high time complexity of the genetic algorithm, we did not carefully set its hyperparameters. Therefore, the experimental results presented in this paper may not represent the optimal performance of our approach.








\bibliography{main}
\clearpage

\appendix
\section{Exploring layer-wise characteristics}
\begin{table}[H]
\centering
\small
\setlength{\tabcolsep}{4pt} 
\renewcommand{\arraystretch}{1.3} 
\begin{tabular}{cccc}
\textbf{Layers operated on} & \textbf{First} & \textbf{Middle} & \textbf{Last}\\
    \hline
    \multicolumn{4}{c}{\textbf{Baseline}} \rule{0pt}{11.5pt} \\
     ——  & 68.33  & 60.30 & 64.33 \\
    \multicolumn{4}{c}{\textbf{2 layers}} \rule{0pt}{11.5pt} \\
        2,3 &  68.00  &  60.33  &  63.67  \\ 
        4,5 &  68.33  &  59.67  &  \cellcolor{darkgreen}64.67  \\ 
        6,7 &  67.00  &  59.33  &  62.67  \\ 
        8,9 &  63.33  &  64.33  & \cellcolor{darkgreen}65.60  \\ 
        9,10 &  65.67  &  61.00  &  64.33  \\ 
        11,12 &  68.67  &  60.00  &  62.67  \\ 
        13,14 &  \cellcolor{darkgreen}70.33  &  63.33  &  \cellcolor{darkgreen}66.33  \\ 
        15,16 &  \cellcolor{darkgreen}69.33  &  57.67  &  63.33  \\ 
        17,18 &  68.67  &  59.33  &  64.33  \\ 
        19,20 &  69.00  &  60.00  &  64.33  \\ 
        21,22 &  \cellcolor{darkgreen}70.33  &  61.33  &  64.33  \\ 
        23,24 &  69.00  &  61.33  &  64.33  \\ 
        25,26 &  68.67  &  61.33  &  64.33  \\ 
        27,28 &  68.33  &  60.67  &  64.67  \\ 
        29,30 &  69.00  &  59.00  &  63.67 \\ 
    \hdashline 
    \multicolumn{4}{c}{\textbf{3 layers}} \rule{0pt}{11.5pt} \\
        2,3,4 & 68.00  & 59.00  & 61.67  \\ 
        5,6,7 & 68.33  & 59.33  & \cellcolor{darkgreen}66.00  \\ 
        8,9,10 & 59.67  & 64.00  & \cellcolor{darkgreen}66.33  \\ 
        10,11,12 & 68.33  & 59.67  & 62.67  \\ 
        13,14,15 & \cellcolor{darkgreen}71.00  & 63.33  & 64.67  \\ 
        16,17,18 & 68.33  & 58.67  & 64.00  \\ 
        19,20,21 & \cellcolor{darkgreen}71.00  & 58.33  & \cellcolor{darkgreen}65.67  \\ 
        22,23,24 & 70.00  & 62.67  & 63.67  \\ 
        25,26,27 & 67.67  & 61.00  & 64.67  \\ 
        28,29,30 & \cellcolor{darkgreen}70.67  & 61.00  & 63.33 \\

    \hdashline 
    \multicolumn{4}{c}{\textbf{4 layers}} \rule{0pt}{11.5pt} \\
        2,3,4,5 & 68.67  & 59.00  & 64.33  \\ 
        6,7,8,9 & 61.00  & 61.33  & \cellcolor{darkgreen}68.33  \\ 
        10,11,12,13 & 65.33  & 58.00  & 61.00  \\ 
        14,15,16,17 & 68.67  & 61.00  & \cellcolor{darkgreen}67.00  \\ 
        18,19,20,21 & \cellcolor{darkgreen}70.67  & 58.67  & 65.33  \\ 
        22,23,24,25 & \cellcolor{darkgreen}71.00  & 62.67  & \cellcolor{darkgreen}66.00  \\ 
        25,26,27,28 & \cellcolor{darkgreen}70.33  & 61.67  & 64.33 \\

\end{tabular}
\caption{The impact of scaling layers at different positions on the results. The three largest values at context ends were highlighted in blue.}
\label{more_part_layer}
\end{table}

We selected the first 300 samples from the MDQA dataset \cite{Liu_Lin_Hewitt_Paranjape_Bevilacqua_Petroni_Liang} and set the scaling factor to \(1.6\). To investigate how scaling layers at different positions impacts the utilization of context at both ends, we applied uniform scaling factors to consecutive layers of \textbf{Vicuna-7B-v1.5} \cite{Berkeley_Cmu_San} at different positions, ensuring that other layers were unaffected. The results show that scaling the earlier layers enhances the model's focus on the latter parts of the context, whereas scaling the later layers shifts the model's attention to the earlier contextual information. These findings suggest that different layers exhibit distinct characteristics, with variations in how they process and attend to different positions within the context.

\section{Dataset}

\begin{table*}[ht]
    \centering
    \small
    \tabcolsep=0.05cm
    \begin{tabular}{|l|l|l|}
    \hline
        Dataset & Description & Metric \\ \hline
        GovReport & Summarization of long reports. & Geometric mean of Rouge-1/2/L scores \\ 
        SummScreenFD & Summarization of TV shows episodes scripts. & Geometric mean of Rouge-1/2/L scores \\
        QMSum & Query-based summarization over meeting transcripts. & Geometric mean of Rouge-1/2/L scores \\
        SQuALITY & Question-focused summarization over stories. & Geometric mean of Rouge-1/2/L scores \\
        Qasper & Question answering over research papers. & F1 score \\
        NarrativeQA & Question answering about entire books and movie scripts. & F1 score \\
        SpaceDigest & Aggregated sentiment classification over 50 hotel reviews from Space. & Exp\_similarity \\ \hline
    \end{tabular}
    \caption{Introduction and evaluation metrics of the sub-datasets under ZeroScroll.}
    \label{zeroscroll_dataset}
\end{table*}

\begin{figure}[ht]
    \centering
    \includegraphics[width=\linewidth]{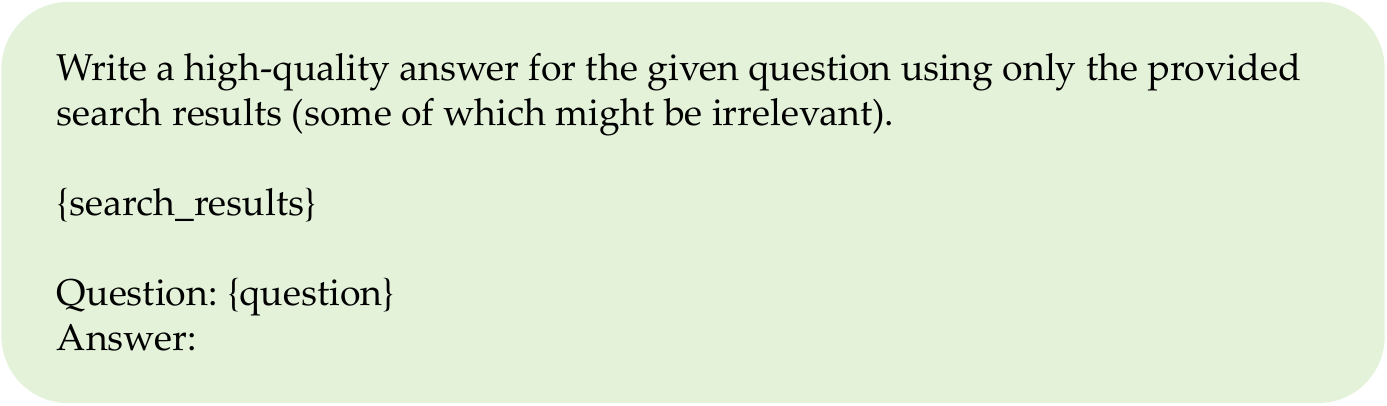}
    \caption{The prompt template of MDQA dataset.}
    \label{MDQA_prompt}
\end{figure}

\begin{figure}[ht]
    \centering
    \includegraphics[width=\linewidth]{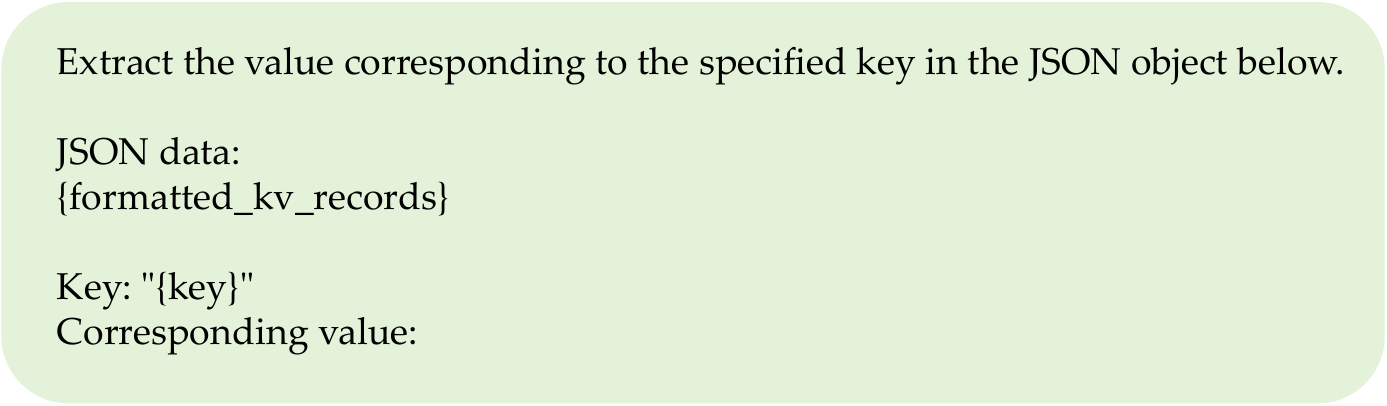}
    \caption{The prompt template of Key-Value Retrieval dataset.}
    \label{KV_prompt}
\end{figure}

We used the MDQA and Key-Value Retrieval datasets \cite{liu2024lost} to investigate the effectiveness of our method in alleviating the "lost-in-the-middle" issue. The corresponding prompt templates for these datasets are shown in Figure \ref{MDQA_prompt} and Figure \ref{KV_prompt}. Building upon these experiments, we further examined whether addressing this issue could enhance the model's ability to leverage contextual information by testing it on the ZeroScroll dataset \cite{Shaham_Ivgi_Efrat_Berant_Levy_2023}. A detailed description of the dataset and the evaluation metrics used can be found in Table \ref{zeroscroll_dataset}. We use the first \(200\) samples of the MDQA dataset to determine the scaling factors and the last \(500\) samples for the ``lost-in-the-middle'' test. For the Key-Value Retrieval dataset, we use \(200\) samples for scaling factors tuning and \(200\) samples for the ``lost-in-the-middle'' test.

\begin{table*}[htp]
    \centering
    \small
    \tabcolsep=0.1cm
    \begin{tabular}{l|l|ccccccc|c}
    \hline
        Model &Method & GovReport & Qasper & SummScreenFd & Qmsum & NarrativeQA & Squality & SpaceDigest & Average \\ \hline
        \multirow{2}{*}{Vicuna-7B-v1.5} & Baseline & 18.44  & 22.82  & 13.71  & 14.50  & 10.98  & 16.56  & 21.39  & 16.91  \\ 
        & Ours & \cellcolor{lightgreen}21.47  & \cellcolor{lightgreen}33.37  & \cellcolor{lightgreen}14.39  & \cellcolor{lightgreen}15.53  & \cellcolor{lightgreen}11.52  & \cellcolor{lightgreen}16.91  & \cellcolor{lightgreen}22.24  & \cellcolor{lightgreen}19.35  \\ \hline
        
        \multirow{2}{*}{LLaMA-2-7B-chat} & Baseline & 18.00  & 13.48  & 13.73  & 14.29  & 10.28  & 15.94  & 49.72  & 19.35  \\ 
        & Ours & \cellcolor{lightgreen}18.20  & \cellcolor{lightgreen}15.23  & \cellcolor{lightgreen}13.99  & \cellcolor{lightgreen}15.04  & \cellcolor{lightgreen}14.93  & \cellcolor{lightgreen}17.37  & \cellcolor{lightgreen}52.28  & \cellcolor{lightgreen}21.01  \\ \hline
        
       \multirow{2}{*}{StableBeluga-7B } & Baseline & 14.88  & 26.89  & 12.09  & 14.24  & 10.73  & 15.05  & 48.50  & 20.34  \\
        & Ours & \cellcolor{lightgreen}18.98  & \cellcolor{lightgreen}34.19  & \cellcolor{lightgreen}13.06  & \cellcolor{lightgreen}15.46  & \cellcolor{lightgreen}9.91  & \cellcolor{lightgreen}16.65  & \cellcolor{lightgreen}46.61  & \cellcolor{lightgreen}22.12 \\
        \hline
    \end{tabular}
    \caption{Our method showed improvements across multiple models on seven long-text tasks, demonstrating its effectiveness in enhancing the model’s ability to utilize contextual information. Descriptions and evaluation metrics of datasets are provided in Table \ref{zeroscroll_dataset}.}
    \label{zeroscroll}
\end{table*}
\newpage
\section{Enhance the utilization of long context}
We further explore the performance of our method in various long-context tasks, as shown in Table \ref{zeroscroll_dataset}.
\section{Discussion Details}

\begin{figure}[ht]
    \centering
    \includegraphics[width=\linewidth]{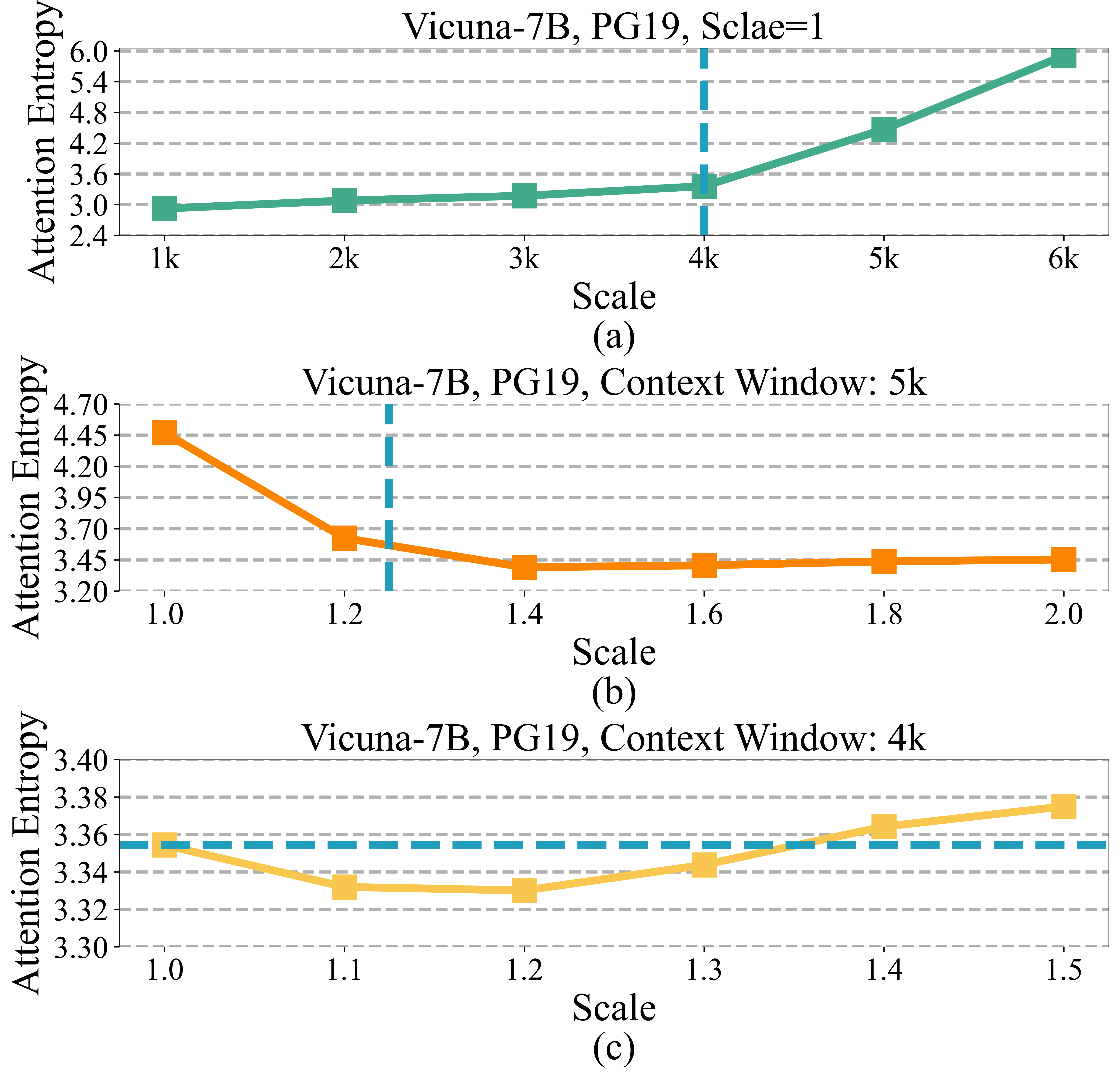}
    \caption{
    Figure \textbf{(a)} shows that as input length exceeds the pre-training context window (4k), attention entropy increases. 
    Figure \textbf{(b)} illustrates that in a 5k window, attention entropy first decreases and then increases with interpolation strength, where the blue line marks the normal strength (5k/4k = 1.25). 
    Figure \textbf{(c)} demonstrates that within the pre-training window, smaller scales concentrate attention while larger scales disperse it.
    }
    \label{disscussion_entropy}
\end{figure}

\begin{figure}[ht]
    \centering
    \includegraphics[width=0.7\linewidth]{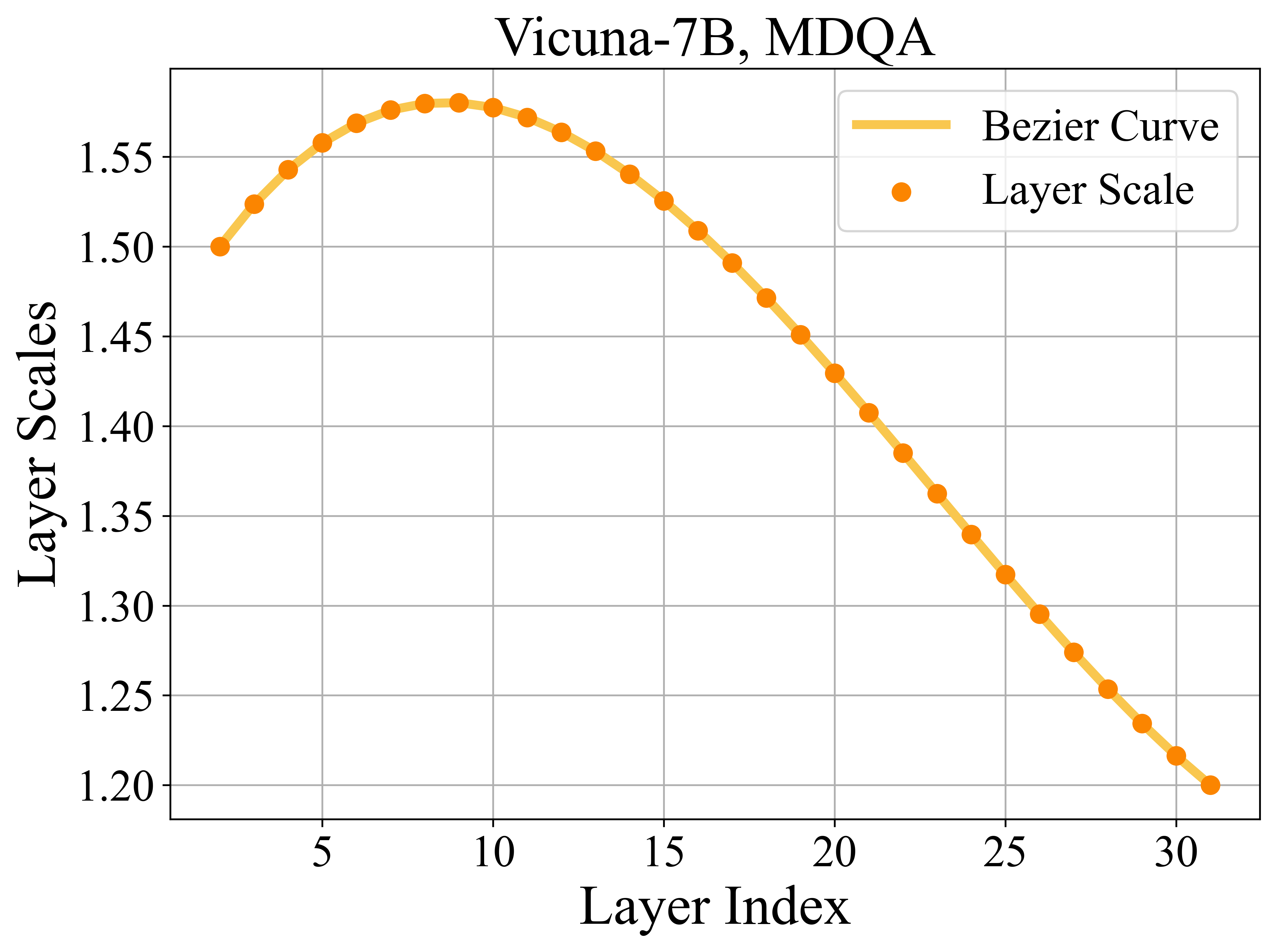}
    \caption{On the MDQA dataset, Bézier curves determined for Vicuna-7B-v1.5 using a genetic algorithm allocated larger scaling factors to the earlier layers and smaller scaling factors to the later layers..}
    \label{B_disscussion_mdqa}
\end{figure}

\begin{figure}[ht]
    \centering
    \includegraphics[width=0.7\linewidth]{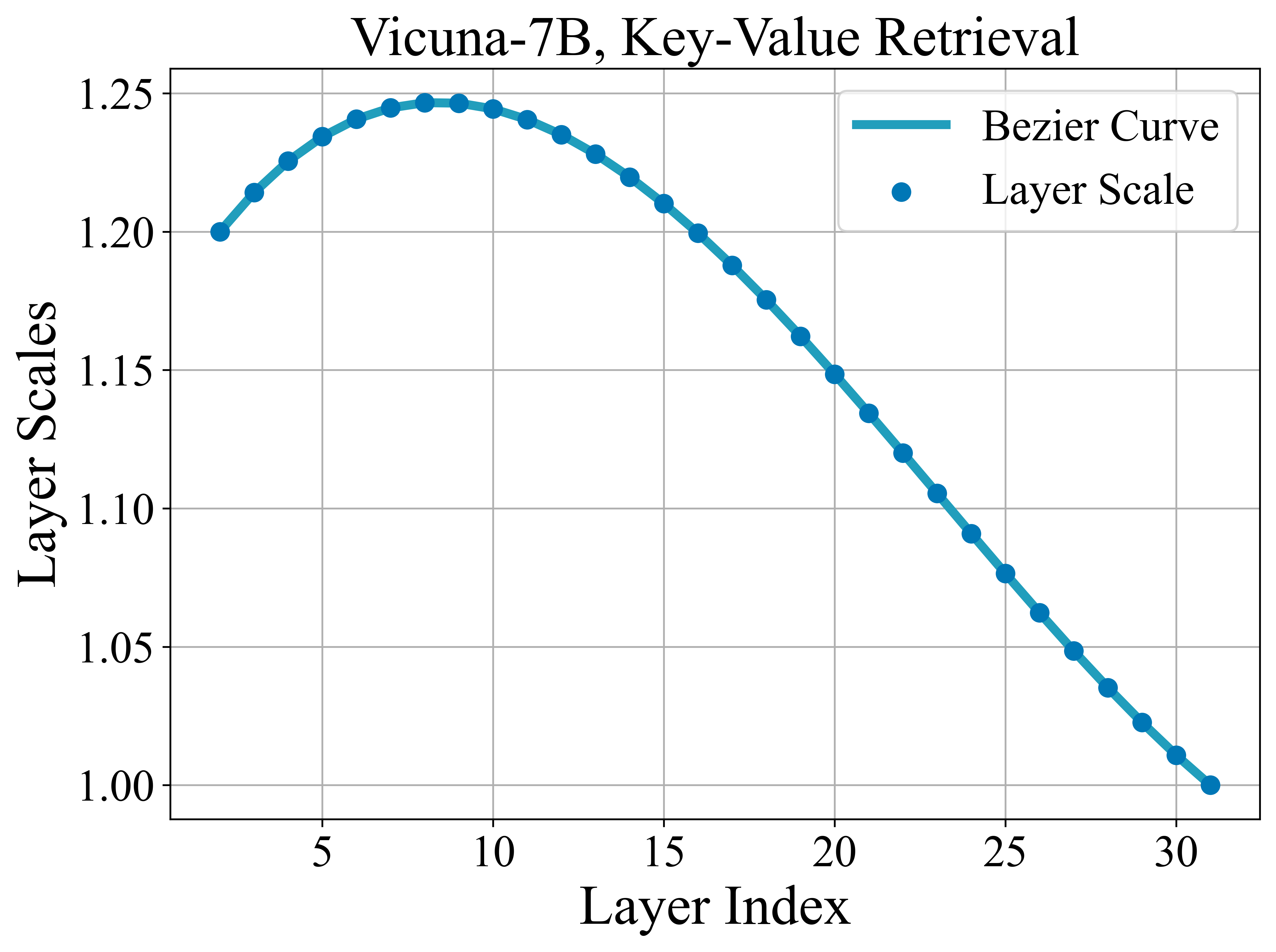}
    \caption{On the Key-Value Retrieval dataset, Bézier curves determined for Vicuna-7B-v1.5 using a genetic algorithm allocated larger scaling factors to the earlier layers and smaller scaling factors to the later layers.}
    \label{B_disscussion_KV}
\end{figure}

\label{sec:appendix}

\end{document}